# Design of Experiments for Calibration of Planar Anthropomorphic Manipulators

Alexandr Klimchik[1,2], Yier Wu[1,2], Stephane Caro[2], Anatol Pashkevich[1,2].

[1]Ecole des Mines de Nantes, 4 rue Alfred-Kastler, Nantes 44307, France
[2]Institut de Recherches en Communications et en Cybernetique de Nantes, 44321 Nantes, France
alexandr.klimchik@mines-nantes.fr, yier.wu@mines-nantes.fr, stephane.caro@irccyn.ec-nantes.fr,
anatol.pashkevich@mines-nantes.fr

**Abstract** – The paper presents a novel technique for the design of optimal calibration experiments for a planar anthropomorphic manipulator with n degrees of freedom. Proposed approach for selection of manipulator configurations allows essentially improving calibration accuracy and reducing parameter identification errors. The results are illustrated by application examples that deal with typical anthropomorphic manipulators.

## I. INTRODUCTION

The standard engineering practice in industrial robotics assumes that the closed-loop control technique is applied only on the level of servo-drives actuating the manipulator joint variables. However, for spatial location of the end-effector, it is applied the open-loop control method that is based on numerous computations of the direct/inverse transformations that define correspondence between the manipulator joint coordinates and the Cartesian coordinates of the end-effector. This requires careful identification (i.e. calibration) of the robot geometric parameters employed in the control algorithm, which usually differ from their nominal values due to manufacturing tolerances [1].

The problem of robot calibration is already well studied and it is in the focus of research community for many years [2]. In general, the calibration process is divided into four sequential steps [3]: modeling, measurements, identification and compensation. First two steps focus on design of the appropriate (complete but non-redundant) mathematical model and carrying out the calibration experiments. Usually, algorithms for the third step are developed for the identification of Denavit-Hartenberg parameters [4], which however are not suitable for the manipulators with collinear axis considered in this paper. For this particular (but very common) case, Hayati [5], Stone [6], and Zhuang [7] proposed some modifications but we will use a more straightforward approach that is more efficient for the planar manipulators.

Among numerous publications devoted to the robot calibration, there is very limited number of works that directly address the issue of the identification accuracy and reduction of the calibration errors [8-16]. It is obviously clear that the calibration accuracy may be improved by increasing the number of experiments (with the factor $1/\sqrt{m}$, where $m$ is the experiments number). Besides, using diverse manipulator configurations for different experiments looks also intuitively promising and perfectly corresponds to some basic ideas of the classical theory [17] that intends using the factors that are distinct as much as possible. However, the classical results are mostly obtained for very specific models (such as linear regression) and can not be applied directly here due to non-linearity of the relevant expressions.

In this paper, the problem of optimal design of the calibration experiments is studied for case if a $n$-link planar manipulator, which does not cover all architectures used in practice but nevertheless allows to derive some very useful analytical expressions and to propose some simple practical rules defining optimal configurations with respect to the calibration accuracy. Particular attention is given to two- and three-link manipulators that are essential components of all existing anthropomorphic robots.

## II. PROBLEM STATEMENT

Let us consider a general $n$-link planar manipulator which geometry can be defined by equations

$$x = \sum_{i=1}^{n}\left[\left(l_i^0 + \Delta l_i\right)\cdot\cos\sum_{j=1}^{i}\left(q_j^0 + \Delta q_j\right)\right]$$
$$y = \sum_{i=1}^{n}\left[\left(l_i^0 + \Delta l_i\right)\cdot\sin\sum_{j=1}^{i}\left(q_j^0 + \Delta q_j\right)\right] \quad (1)$$

where $(x, y)$ is the end-effector position, $l_i^0$, $q_j^0$ are the nominal length and angular coordinates of the $i$-th link and actuator respectively, $\Delta l_i$ and $\Delta q_i$ are their deviations from nominal values, $n$ is the number of links. Let us also introduce notations $\theta_i^0 = \sum_{j=1}^{i} q_j^0$, and $\Delta\theta_i = \sum_{j=1}^{i} \Delta q_j$ that will be useful for further computations. As follows from (1), the manipulators geometrical model includes $2n$ parameters $\{\Delta l_i, \Delta\theta_i, i = \overline{1,n}\}$ that must be identified by means of the calibration.

It is assumed that each calibration experiment produces two vectors, which define the Cartesian coordinates of the end-effector $\mathbf{P}_i = [x_i \ y_i]^T$ and corresponding joint coordinates $\mathbf{Q}_i = (q_{1i}, q_{2i}, \ldots, q_{ni})$. Besides, the measurement errors for the Cartesian coordinates $(\varepsilon_x, \varepsilon_y)$ are assumed to be iid (independent identically distributed) random values with zero mean and standard deviation $\sigma$, while the measurement errors for the joint variables are relatively small. Hence, the calibration procedure may be treated as the best fitting of the experimental data $\{\mathbf{Q}_i, \mathbf{P}_i\}$ by using the geometrical model (1) that leads to the standard least-square problem. However, due to the errors in the measurements, the desired values $\{\Delta l_i, \Delta \theta_i, i = \overline{1,n}\}$ are always identified approximately. So, the problem of interest is to evaluate (in the frame of the above assumption) the identification accuracy for the parameters $\{\Delta l_i, \Delta \theta_i, i = \overline{1,n}\}$ and to propose a technique for selecting the set of the joint variables $\mathbf{Q}_i = (q_{1i}, q_{2i}, \ldots, q_{ni})$ that leads to improvement of this accuracy (in statistical sense).

To solve this general problem, let us sequentially present the calibration algorithm, evaluate related identification errors and develop optimality conditions allowing minimize the number of experiments for given accuracy in identification of the desired parameters.

### III. CALIBRATION ALGORITHM

As follows from the previous Section, the input data for the manipulator calibration are its joint coordinates $\mathbf{Q}_i = (q_{1i}, q_{2i}, \ldots, q_{ni})$ and corresponding end-effector positions $\mathbf{P}_i = [x_i \ y_i]^T$, $i = \overline{1,m}$. The goal is to find unknown parameters $\Delta\mathbf{\Pi} = \{\Delta l_i, \Delta \theta_i, i = \overline{1,n}\}$ which ensure the best mapping of the coordinates $\mathbf{Q}_i$ to the end-effector positions $\mathbf{P}_i$ that is defined by the geometrical model (1), which may be re-written in a general form as

$$x_i = f_x(\mathbf{Q}_i, \Delta\mathbf{\Pi}); \quad y_i = f_y(\mathbf{Q}_i, \Delta\mathbf{\Pi}); \quad i = \overline{1,m} \quad (2)$$

where $f_x(\mathbf{Q}_i, \Delta\mathbf{\Pi})$, $f_y(\mathbf{Q}_i, \Delta\mathbf{\Pi})$ are the right-hand sides of system (1).

To compute $\Delta\mathbf{\Pi} = \{\Delta l_i, \Delta \theta_i, i = \overline{1,n}\}$, let us apply the least-square method which minimizes the residuals for all experimental configurations. Corresponding optimization problem can be written as

$$F = \sum_{i=1}^{m} \left( \left(f_x(\mathbf{Q}_i, \Delta\mathbf{\Pi}) - x_i\right)^2 + \left(f_y(\mathbf{Q}_i, \Delta\mathbf{\Pi}) - y_i\right)^2 \right) \to \min \quad (3)$$

and it can be solved by using stationary condition at the extreme point $\partial F / \partial \mathbf{\Pi}_i = 0$ for $i = \overline{1,2n}$ with respect to $\Delta\mathbf{\Pi} = \{\Delta l_i, \Delta \theta_i, i = \overline{1,n}\}$. Corresponding derivations yield

$$\frac{\partial F}{\partial \Delta \theta_k} = \sum_{i=1}^{m} \left[ -\sum_{l=i}^{n} \left( (l_l^0 + \Delta l_l) \sin\left(\theta_l^{0(i)} + \Delta \theta_l\right) \right) \times \right.$$
$$\times \left( \sum_{l=1}^{n} \left( (l_l^0 + \Delta l_l) \cos\left(\theta_l^{0(i)} + \Delta \theta_l\right) \right) - x_i \right) +$$
$$+ \sum_{l=i}^{n} \left( (l_l^0 + \Delta l_l) \cos\left(\theta_l^{0(i)} + \Delta \theta_l\right) \right) \times$$
$$\left. \times \left( \sum_{l=1}^{n} \left( (l_l^0 + \Delta l_l) \sin\left(\theta_l^{0(i)} + \Delta \theta_l\right) \right) - y_i \right) \right] = 0 \quad (4)$$

$$\frac{\partial F}{\partial \Delta l_k} = \sum_{i=1}^{m} \left[ (\cos(\theta_k^{0(i)} + \Delta \theta_k) \times \right.$$
$$\times \left( \sum_{l=1}^{n} \left( (l_l^0 + \Delta l_l) \cos(\theta_l^{0(i)} + \Delta \theta_l) \right) - x_i \right) + \sin(\theta_k^{0(i)} + \Delta \theta_k) \times$$
$$\left. \times \left( \sum_{l=1}^{n} \left( (l_l^0 + \Delta l_l) \sin(\theta_l^{0(i)} + \Delta \theta_l) \right) - y_i \right) \right] = 0 \quad (5)$$

where $k = \overline{1,n}$, $\theta_j^{(i)} = \sum_{k=1}^{j} q_k^{(i)}$ is the orientation of j-th link in the i-th experiment. Since this system of equations is nonlinear with respect to $\Delta \theta_i$, it does not have general analytical solution. Thus, it is reasonable to linearize the model (1)

$$\mathbf{P}_i = \mathbf{P}_{0i} + \mathbf{J}_i \cdot \Delta\mathbf{\Pi} \quad (6)$$

where $\mathbf{P}_{0i}$ is the end-effector position for the nominal values of parameters and the joint variables

$$\mathbf{P}_{0i} = \left[ \sum_{k=1}^{n} l_k^0 \cos\theta_k^{0(i)} \quad \sum_{k=1}^{n} l_k^0 \sin\theta_k^{0(i)} \right]^T, \quad i = \overline{1,m},$$

$\mathbf{J}_i$ is the Jacobian matrix, which can be computed by differencing the system (1) with respect to $\Delta\mathbf{\Pi}$ that leads to

$$\mathbf{J}_i = \begin{bmatrix} \mathbf{J}_{xq}^{(i)} & \mathbf{J}_{xl}^{(i)} \\ \mathbf{J}_{yq}^{(i)} & \mathbf{J}_{yl}^{(i)} \end{bmatrix}_{2 \times 2n} \quad (7)$$

where

$$\mathbf{J}_{xq}^{(i)} = \left[ -l_1 \cdot \sin\theta_1^{(i)} \quad \ldots \quad -l_n \cdot \sin\theta_n^{(i)} \right]_{1 \times n}$$
$$\mathbf{J}_{yq}^{(i)} = \left[ l_1 \cdot \cos\theta_1^{(i)} \quad \ldots \quad l_j \cdot \cos\theta_j^{(i)} \right]_{1 \times n}$$
$$\mathbf{J}_{xl}^{(i)} = \left[ \cos\theta_1^{(i)} \quad \ldots \quad \cos\theta_n^{(i)} \right]_{1 \times n} \quad (8)$$
$$\mathbf{J}_{yl}^{(i)} = \left[ \sin\theta_1^{(i)} \quad \ldots \quad \sin\theta_n^{(i)} \right]_{1 \times n}$$

Taking into account (6), the function (3) can be rewritten as

$$F = \sum_{i=1}^{m} \left( (\mathbf{J}_i \cdot \Delta \mathbf{\Pi} - \Delta \mathbf{P}_i)^T (\mathbf{J}_i \cdot \Delta \mathbf{\Pi} - \Delta \mathbf{P}_i) \right) \to \min \quad (9)$$

where $\Delta \mathbf{P}_i = \mathbf{P}_i - \mathbf{P}_{0i}$ and expressions (4) are reduced to

$$\sum_{i=1}^{m} (\mathbf{J}_i^T \cdot \mathbf{J}_i) \cdot \Delta \mathbf{\Pi} = \sum_{i=1}^{m} (\mathbf{J}_i^T \cdot \Delta \mathbf{P}_i). \quad (10)$$

So, the unknown parameters $\Delta \mathbf{\Pi}$, can be computed as

$$\Delta \mathbf{\Pi} = (\mathbf{J}_a^T \cdot \mathbf{J}_a)^{-1} \cdot \mathbf{J}_a^T \cdot \Delta \mathbf{P}_a \quad (11)$$

where $\mathbf{J}_a = [\mathbf{J}_1 \ \mathbf{J}_2 \ ... \ \mathbf{J}_m]^T$; $\Delta \mathbf{P}_a = [\Delta \mathbf{P}_1 \ \Delta \mathbf{P}_2 \ ... \ \Delta \mathbf{P}_m]^T$.

To increase the identification accuracy, the foregoing linearized procedure has to be applied several times, in accordance with the following iterative algorithm:

**Step 1.** Carry out experiments and collect the input data in the vectors of generalized coordinates $\mathbf{Q}_i$ and end-effector position $\mathbf{P}_i(x_i, y_i)$. Initialize $\Delta \mathbf{\Pi} = 0$.

**Step 2.** Compute end-effector position via direct kinematic model (1) using initial generalized coordinates $\mathbf{Q}_i$

**Step 3.** Compute residuals and unknown parameters $\Delta \mathbf{\Pi}$ via (11)

**Step 4.** Correct mathematical model and generalized coordinates $l_j = l_j + \Delta l_j$, $\theta_{ji} = \theta_{ji} + \Delta \theta_j$, $j = \overline{1, m}$.

**Step 5.** If required accuracy is not satisfied, repeat from Step 2.

It should be mentioned, that the proposed iterative algorithm can produce exact values of $\{\Delta l_i, \Delta \theta_i, i = \overline{1, n}\}$ if and only if there are no measurement errors in the initial data $\{\mathbf{Q}_i, \mathbf{P}_i\}$. Since in practice it is not true, it is reasonable to minimize the measurement errors impact via proper selection of $\{\mathbf{Q}_i, \mathbf{P}_i\}$.

## IV. ACCURACY OF CALIBRATION EXPERIMENT

Let us assume that the measurements of *x*, *y* are carrying out with some random errors $\varepsilon_{xi}$, $\varepsilon_{yi}$ that are assumed to be iid, with the standard deviation $\sigma$ and zero mean value. Thus, model (6) can be rewritten as

$$\mathbf{P}_i = \mathbf{P}_{0i} + \mathbf{J}_i \cdot \Delta \mathbf{\Pi} + \boldsymbol{\varepsilon}_i \quad (12)$$

where the vector $\boldsymbol{\varepsilon}_i = [\varepsilon_{xi}, \varepsilon_{yi}]^T$ collect all measurement errors. So, expression (11) for computing the vector of the desired parameters $\Delta \mathbf{\Pi}$ has to be rewritten as

$$\Delta \mathbf{\Pi} = (\mathbf{J}_a^T \cdot \mathbf{J}_a)^{-1} \cdot \mathbf{J}_a^T \cdot (\Delta \mathbf{P}_a - \boldsymbol{\varepsilon}_a) \quad (13)$$

where $\boldsymbol{\varepsilon}_a = [\boldsymbol{\varepsilon}_1 \ \boldsymbol{\varepsilon}_2 \ ... \ \boldsymbol{\varepsilon}_m]^T$.

As follows from (13), the latter expression produces unbiased estimates

$$E(\Delta \mathbf{\Pi}) = (\mathbf{J}_a^T \cdot \mathbf{J}_a)^{-1} \cdot \mathbf{J}_a^T \cdot \Delta \mathbf{P}_a \quad (14)$$

Besides, it can be proved that the covariance matrix of the parameters $\Delta \mathbf{\Pi}$ [18], defining the identification accuracy, can be expressed as

$$\mathrm{cov}(\Delta \mathbf{\Pi}) = (\mathbf{J}_a^T \cdot \mathbf{J}_a)^{-1} \cdot \mathbf{J}_a^T \cdot E(\boldsymbol{\varepsilon}_a \boldsymbol{\varepsilon}_a^T) \cdot \mathbf{J}_a \cdot (\mathbf{J}_a^T \cdot \mathbf{J}_a)^{-1} \quad (15)$$

Then, taking into account that $E(\boldsymbol{\varepsilon}_a \boldsymbol{\varepsilon}_a^T) = \sigma^2 \mathbf{I}_{2n}$, where $\mathbf{I}_{2n}$ is the identity matrix of the size $2n \times 2n$, the expression (15) can be simplified to

$$\mathrm{cov}(\Delta \mathbf{\Pi}) = \sigma^2 \left( \sum_{i=1}^{m} (\mathbf{J}_i^T \cdot \mathbf{J}_i) \right)^{-1} \quad (16)$$

Therefore, for the problem of interest, the impact of the measurement errors (i.e. "quality" of the experiment plan) is defined by the matrix sum $\sum_{i=1}^{m} (\mathbf{J}_i^T \cdot \mathbf{J}_i)$.

For the considered model (1), this sum can be expressed as

$$\sum_{i=1}^{m} (\mathbf{J}_i^T \cdot \mathbf{J}_i) = \begin{bmatrix} \mathbf{A} & \mathbf{B} \\ \mathbf{C} & \mathbf{D} \end{bmatrix} \quad (17)$$

where

$$\mathbf{A} = [l_j l_k \cdot c_{jk}]; \quad \mathbf{B} = \mathbf{C} = [l_j s_{jk}]; \quad \mathbf{D} = [c_{jk}];$$

$$c_{jk} = \sum_{i=1}^{m} \cos(\theta_j^{(i)} - \theta_k^{(i)}); \quad s_{jk} = \sum_{i=1}^{m} \sin(\theta_j^{(i)} - \theta_k^{(i)}); \quad (18)$$

$$j = \overline{1, n}; \ k = \overline{1, n}$$

where $A_{j,j} = m \cdot l_j l_k$; $B_{j,j} = C_{j,j} = 0$; $D_{j,j} = m$, $j = \overline{1, n}$, which can be presented via block matrix

$$\sum_{i=1}^{m} (\mathbf{J}_i^T \cdot \mathbf{J}_i) = \begin{bmatrix} \mathbf{L} \cdot \mathbf{C} \cdot \mathbf{L} & \mathbf{L} \cdot \mathbf{S} \\ (\mathbf{L} \cdot \mathbf{S})^T & \mathbf{C} \end{bmatrix} \quad (19)$$

where $\mathbf{L} = diag(l_1, l_2, ..., l_n)$, $\mathbf{C} = [c_{jk}]$, $j = \overline{1, n}; \ k = \overline{1, n}$, $\mathbf{S} = [s_{jk}]$, $j = \overline{1, n}; \ k = \overline{1, n}$.

This expression allows estimating the identification accuracy and it can be applied for optimal design of calibration that is presented in the following Section.

## V. DESIGN OF CALIBRATION EXPERIMENTS

To optimize location of experimental points in the Cartesian space (and corresponding manipulator configurations), let us investigate in details all components of the matrix $\sum_{i=1}^{m}\left(\mathbf{J}_i^T \cdot \mathbf{J}_i\right)$ that is similar to the "information matrix" in classical design of experiments. As it is known [17], this matrix can be evaluated by several criteria. The most common of them are A- and D-optimality criteria, but here it is not reasonable to use the A- criterion because the trace of the matrix $\sum_{i=1}^{m}\left(\mathbf{J}_i^T \cdot \mathbf{J}_i\right)$ does not depend on the experiment plan. Besides, the D- criterion is also not applicable here in its direct form.

Hence, let us introduce a modified D*-optimal criterion which takes into account the structure of the information matrix in this particular case. Since this matrix includes several blocks with different units (linear, angular, etc.), it is reasonable to focus on optimization of each block separately. This approach allows to reformulate the problem and to define the goal as

$$\left|\det(\mathbf{C}')\right| \to \max_{q_j,\ j=\overline{1,m}}; \quad \left|\det(\mathbf{S}')\right| \to \min_{q_j,\ j=\overline{1,m}} \quad (20)$$

where $\mathbf{C}' = [c_{jk}/n]$, $\mathbf{S}' = [s_{jk}/n]$ correspond to the diagonal and non-diagonal blocks of (19) respectively. It can be proved that this goal is satisfied if

$$c_{jk} = 0; \quad s_{jk} = 0; \quad j = \overline{1,n};\ k = \overline{1,n};\ j \neq k \quad (21)$$

that perfectly corresponds to the classical D-optimality conditions. For practical convenience, cases of 2-, 3- and 4-links manipulators were investigated in details and corresponding optimality conditions are presented in Table 1.

A correspondence between the proposed approach and the D-optimality can be also proved analytically. In particular, straightforward computations give

$$\frac{\partial c_{jk}}{\partial \theta_b} = \begin{cases} 0, & \text{if } b \neq j, b \neq k \\ -s_{jk}, & \text{if } b = j \\ s_{jk}, & \text{if } b = k \end{cases} ; \quad \frac{\partial s_{jk}}{\partial \theta_b} = \begin{cases} 0, & \text{if } b \neq j, b \neq k \\ c_{jk}, & \text{if } b = j \\ -c_{jk}, & \text{if } b = k \end{cases} \quad (22)$$

which leads to

$$\partial \mathbf{C} / \partial \theta_b = 0; \quad \partial \mathbf{S} / \partial \theta_b = 0. \quad (23)$$

The latter guarantees maximum of the relevant determinant and ensures agreement with the D-optimality.

Validity of the proposed approach and its practical significance was also conformed by a simulation example that deals with 4-links manipulator with geometrical parameters $l_1 = 260\,mm$, $l_2 = 180\,mm$, $l_3 = 120\,mm$, $l_4 = 100\,mm$ and their deviations $\Delta l_1 = 1.5\,mm$, $\Delta l_2 = -0.6\,mm$, $\Delta l_3 = -0.4\,mm$, $\Delta l_4 = 0.7\,mm$; and deviation of zero values of angular coordinates $\Delta q_1 = 0.5°$, $\Delta q_2 = -0.5°$, $\Delta q_3 = 0.7°$, $\Delta q_4 = -0.3°$. All experiments were carried out for 10 random experimental points, the results are summarized in the Figure 1. They show that random plans give rather poor results both for D-optimality and D*-optimality criteria comparing to the optimal ones (for the optimal plans $\det(\mathbf{C}') = 1$ and $\det(\mathbf{S}') = 0$; $\det(\mathbf{D}') = 1$, where $\mathbf{D}'$ is normalized block matrix (19)).

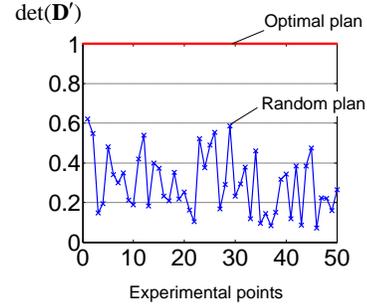

Figure 1. Determinant values of matrix $\mathbf{D}'$ for 4-links manipulator for random calibration plans with 10 experimental points:

TABLE I. OPTIMAL PLAN CONDITIONS FOR 2-, 3- AND 4-LINKS MANIPULATORS

| Manipulator | Conditions for optimal plan | Notation |
|---|---|---|
| 2-links manipulator | $\sum_{i=1}^{m} c_{2i} = 0;\ \sum_{i=1}^{m} s_{2i} = 0$, | $c_{2i} = \cos q_{2i}$; $s_{2i} = \sin q_{2i}$; $i = \overline{1,m}$ |
| 3-links manipulator | $\sum_{i=1}^{m} c_{2i} = 0;\ \sum_{i=1}^{m} s_{2i} = 0;\ \sum_{i=1}^{m} c_{3i} = 0;\ \sum_{i=1}^{m} s_{3i} = 0;\ \sum_{i=1}^{m} c_{23i} = 0;\ \sum_{i=1}^{m} s_{23i} = 0$ | $c_{3i} = \cos q_{3i}$; $s_{3i} = \sin q_{3i}$; $c_{23i} = \cos(q_{2i}+q_{3i})$; $s_{3i} = \sin(q_{2i}+q_{3i})$; $i = \overline{1,m}$ |
| 4-links manipulator | $\sum_{i=1}^{m} c_{2i} = 0;\ \sum_{i=1}^{m} s_{2i} = 0;\ \sum_{i=1}^{m} c_{3i} = 0;\ \sum_{i=1}^{m} s_{3i} = 0;\ \sum_{i=1}^{m} c_{4i} = 0;\ \sum_{i=1}^{m} s_{4i} = 0;$ $\sum_{i=1}^{m} c_{23i} = 0;\ \sum_{i=1}^{m} s_{23i} = 0;\ \sum_{i=1}^{m} c_{24i} = 0;\ \sum_{i=1}^{m} s_{24i} = 0;\ \sum_{i=1}^{m} c_{34i} = 0;\ \sum_{i=1}^{m} s_{34i} = 0$ | $c_{4i} = \cos q_{4i}$; $s_{4i} = \sin q_{4i}$; $c_{24i} = \cos(q_{2i}+q_{3i}+q_{4i})$; $s_{24i} = \sin(q_{2i}+q_{3i}+q_{4i})$; $c_{34i} = \cos(q_{3i}+q_{4i})$; $s_{34i} = \sin(q_{3i}+q_{4i})$; $i = \overline{1,m}$ |

For the proposed set of calibration experiments, the calibration accuracy can be estimated via the covariance matrix, which in this case is diagonal and may be presented as

$$\operatorname{cov}(\Delta \mathbf{\Pi}) = \sigma^2 \cdot \begin{bmatrix} m \cdot \mathbf{L} \cdot \mathbf{L} & \mathbf{0} \\ \mathbf{0} & \mathbf{I} \end{bmatrix} \quad (24)$$

where $\mathbf{L} = diag(l_1, l_2, ..., l_n)$, and identification accuracy can be evaluated as

$$\sigma_{qi} = \frac{\sigma}{\sqrt{m} \cdot l_i}; \qquad \sigma_{Li} = \frac{\sigma}{\sqrt{m}}; \qquad i = \overline{1, n} \quad (25)$$

where $\sigma_{qi}, \sigma_{Li}$ are standard deviations of angular ($q_i$) and linear ($l_i$) parameters from the nominal values.

The results show that identification errors of the linear parameters depend only on the number of experimental points, while the angular parameter errors also depend on the link length.

## VI. SIMULATION STUDY

Let us present some simulation results that demonstrate efficiency of the proposed technique for several case studies that deal with two-, three- and four-links manipulators and employ different number of calibration experiments. It is assumed that in all cases the calibration experiments were designed in accordance with expressions developed in Section 5 (see Table 1). To obtain meaningful statistics, the simulation was repeated 10000 times; the deviation of measurement error $\sigma$ was equal to 0.1 mm.

It was also assumed that the manipulator geometrical parameters are $l_1 = 260\,mm$, $l_2 = 180\,mm$, $l_3 = 120\,mm$, $l_4 = 100\,mm$ and their deviations are equal to 1.5 mm, -0.6 mm, -0.4 mm and 0.7 mm. respectively, while the deviation of zero values of angular coordinates 0.5°, -0.5°, 0.7° and -0.3° for the first, second, third and fourth joints respectively. Short summary of the simulation results are presented in Table 3 and in Figure 2.

As follows from this study, the identification accuracy of the experimental result and analytical estimations are in good agreement. In particular, for linear parameters, the identification error reduces from 0.022 mm to 0.005 mm while the experiment number increases from 4 to 20. Besides, these results allow defining minimum number of experimental points to satisfy the required accuracy. Thus, to satisfy an accuracy of 0.001 mm for linear parameters it is required to carry out 100 experiments, which will provide accuracy for angular parameters 0.002°, 0.003°, 0.005° and 0.006° respectively.

TABLE II.  ESTIMATION OF THE IDENTIFICATION ACCURACY OF GEOMETRICAL PARAMETERS: ANALYTICAL SOLUTION

| Manipulator | $\sum_{i=1}^{m}(J^T_i \cdot J_i)$ | Identification accuracy |
|---|---|---|
| 2-links manipulator | $diag(m \cdot l_1^2,\ m \cdot l_2^2,\ m,\ m)$ | $\sigma_{q1} = \frac{\sigma}{\sqrt{m} \cdot l_1};\ \sigma_{q2} = \frac{\sigma}{\sqrt{m} \cdot l_2};\ \sigma_{L1} = \frac{\sigma}{\sqrt{m}};\ \sigma_{L2} = \frac{\sigma}{\sqrt{m}}$ |
| 3-links manipulator | $diag(m \cdot l_1^2,\ m \cdot l_2^2,\ m \cdot l_3^2,\ m,\ m,\ m)$ | $\sigma_{q1} = \frac{\sigma}{\sqrt{m} \cdot l_1};\ \sigma_{q2} = \frac{\sigma}{\sqrt{m} \cdot l_2};\ \sigma_{q3} = \frac{\sigma}{\sqrt{m} \cdot l_3};\ \sigma_{L1} = \frac{\sigma}{\sqrt{m}};\ \sigma_{L2} = \frac{\sigma}{\sqrt{m}};\ \sigma_{L3} = \frac{\sigma}{\sqrt{m}}$ |
| 4-links manipulator | $diag(m \cdot l_1^2,\ m \cdot l_2^2,\ m \cdot l_3^2,\ m \cdot l_4^2,\ m,\ m,\ m,\ m)$ | $\sigma_{q1} = \frac{\sigma}{\sqrt{m} \cdot l_1};\ \sigma_{q2} = \frac{\sigma}{\sqrt{m} \cdot l_2};\ \sigma_{q3} = \frac{\sigma}{\sqrt{m} \cdot l_3};\ \sigma_{q4} = \frac{\sigma}{\sqrt{m} \cdot l_4};$ <br> $\sigma_{L1} = \frac{\sigma}{\sqrt{m}};\ \sigma_{L2} = \frac{\sigma}{\sqrt{m}};\ \sigma_{L3} = \frac{\sigma}{\sqrt{m}};\ \sigma_{L4} = \frac{\sigma}{\sqrt{m}}$ |

TABLE III.  ESTIMATION OF IDENTIFICATION ACCURACY OF GEOMETRICAL PARAMETERS

| Manipulator | Model parameters | Identification accuracy | |
|---|---|---|---|
| | | 3 experimental points | 20 experimental points |
| 2-links manipulator | $L_1 = 260\,mm,\ \Delta L_1 = 1.5\,mm,\ \Delta q_1 = 0.5\,deg$ <br> $L_2 = 180\,mm,\ \Delta L_2 = -0.6\,mm,\ \Delta q_2 = -0.5\,deg$ | $\Delta L_1 = 0.058\,mm,\ \Delta q_1 = 0.013\,deg$ <br> $\Delta L_2 = 0.058\,mm,\ \Delta q_2 = 0.018\,deg$ | $\Delta L_1 = 0.058\,mm,\ \Delta q_1 = 0.005\,deg$ <br> $\Delta L_2 = 0.058\,mm,\ \Delta q_2 = 0.007\,deg$ |
| 3-links manipulator | $L_1 = 260\,mm,\ \Delta L_1 = 1.5\,mm,\ \Delta q_1 = 0.5\,deg$ <br> $L_2 = 180\,mm,\ \Delta L_2 = -0.6\,mm,\ \Delta q_2 = -0.5\,deg$ <br> $L_3 = 120\,mm,\ \Delta L_3 = -0.4\,mm,\ \Delta q_3 = 0.7\,deg$ | $\Delta L_1 = 0.058\,mm,\ \Delta q_1 = 0.013\,deg$ <br> $\Delta L_2 = 0.058\,mm,\ \Delta q_2 = 0.018\,deg$ <br> $\Delta L_3 = 0.058\,mm\ \ \Delta q_3 = 0.027\,deg$ | $\Delta L_1 = 0.022\,mm,\ \Delta q_1 = 0.005\,deg$ <br> $\Delta L_2 = 0.022\,mm,\ \Delta q_2 = 0.007\,deg$ <br> $\Delta L_3 = 0.022\,mm\ \ \Delta q_3 = 0.011\,deg$ |
| 4-links manipulator | $L_1 = 260\,mm,\ \Delta L_1 = 1.5\,mm,\ \Delta q_1 = 0.5\,deg$ <br> $L_2 = 180\,mm,\ \Delta L_2 = -0.6\,mm,\ \Delta q_2 = -0.5\,deg$ <br> $L_3 = 120\,mm,\ \Delta L_3 = -0.4\,mm,\ \Delta q_3 = 0.7\,deg$ <br> $L_4 = 100\,mm,\ \Delta L_4 = 0.7\,mm,\ \Delta q_4 = -0.3\,deg$ | $\Delta L_1 = 0.058\,mm,\ \Delta q_1 = 0.013\,deg$ <br> $\Delta L_2 = 0.058\,mm,\ \Delta q_2 = 0.018\,deg$ <br> $\Delta L_3 = 0.058\,mm\ \ \Delta q_3 = 0.027\,deg$ <br> $\Delta L_4 = 0.058\,mm\ \ \Delta q_4 = 0.033\,deg$ | $\Delta L_1 = 0.022\,mm,\ \Delta q_1 = 0.005\,deg$ <br> $\Delta L_2 = 0.022\,mm,\ \Delta q_2 = 0.007\,deg$ <br> $\Delta L_3 = 0.022\,mm\ \ \Delta q_3 = 0.011\,deg$ <br> $\Delta L_4 = 0.022\,mm\ \ \Delta q_4 = 0.013\,deg$ |

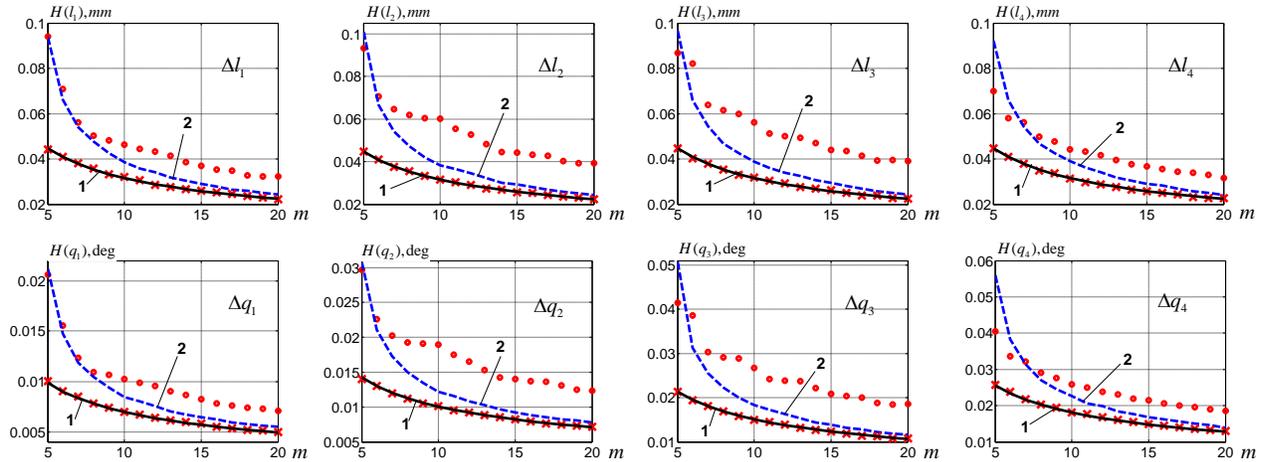

Figure 2.  Identification accuracy for the geometrical parameters identification of 4-links manipulator with optimal experiment planning:
"x" are experimental values corresponding to the optimal calibration plan, "o" are experimental values corresponding to the standard calibration plan
"1" is analytical curve coresponding to the optimal plan, "2" is an average experemental curve corresponding to 10000 random calibtration plans.

## VII. CONCLUSION

The paper presents a new approach for design of calibration experiments that allows essentially reducing the identification errors due to proper selection of the manipulator postures employed in the measurements. There were obtained analytical expressions describing set of the optimal postures corresponding the proposed D*-criterion that is adopted to special structure of the information matrix. Validity of the obtained results and their practical significance were confirmed via simulation study that deals with two-, three- and four-links planar manipulators.

Compared to previous contributions, these results can be treated as further development of the design-of-experiments theory that is adapted to the specific type of the non-linear models that arise in robot kinematics. Future work will focus on extension of these results for non-planar manipulators.

ACKNOWLEDGEMENTS

The work presented in this paper was partially funded by the Region "Pays de la Loire", France and by the project ANR COROUSSO, France.